\documentclass[lettersize,journal]{IEEEtran}
\usepackage{amsmath,amsfonts}
\usepackage{algorithmic}
\usepackage{algorithm}
\usepackage{array}
\usepackage[caption=false,font=normalsize,labelfont=sf,textfont=sf]{subfig}
\usepackage{textcomp}
\usepackage{booktabs} 
\usepackage{multirow}
\usepackage{stfloats}
\usepackage{url}
\usepackage{verbatim}
\usepackage{graphicx}
\usepackage{cite}
\usepackage{bm}
\usepackage{makecell}
\usepackage{bbding}

\begin{document}

\title{Rethinking Supervision in Document Unwarping: A Self-consistent Flow-free Approach}

\author{Shaokai Liu$^{\dagger}$,
Hao Feng$^{\dagger}$,
Wengang Zhou*

\IEEEcompsocitemizethanks{
\IEEEcompsocthanksitem This work is supported by National Natural Science Foundation of China under Contract 62021001. It was also supported by GPU cluster built by MCC Lab of Information Science and Technology Institution, USTC, and the Supercomputing Center of the USTC.
\IEEEcompsocthanksitem Shaokai Liu, Hao Feng, and Wengang Zhou are with MoE Key Laboratory of Brain-inspired Intelligent Perception and Cognition, University of Science and Technology of China, Hefei, 230027, China.
E-mail: \{liushaokai, haof\}@mail.ustc.edu.cn; zhwg@ustc.edu.cn.
\IEEEcompsocthanksitem  $^{\dagger}$The first two authors contribute equally to this work.
\IEEEcompsocthanksitem  *Corresponding author: Wengang Zhou.

}}

\maketitle

\begin{abstract}
In recent years, the proliferation of smartphones has led to an upsurge in the digitization of document files via these portable devices.
However, images captured by smartphones often suffer from distortions, thereby negatively affecting digital preservation and downstream applications. 
To address this issue, we introduce DRNet, a novel deep network for document image rectification. 
Our approach is based on three key designs. 
Firstly, we exploit the intrinsic geometric consistency inherent in document images to guide the learning process of distortion rectification. Secondly, we design a coarse-to-fine rectification network to leverage the representations derived from the distorted document image, thereby enhancing the rectification result.
Thirdly,
we propose a unique perspective for supervising the learning of rectification networks, where undistorted document images are employed for supervision,
which is free of warping mesh as ground truth in existing methods.
Technically, both low-level pixel alignment and high-level
semantic alignment jointly contribute to the learning of the mapping relationship between deformed document images and
distortion-free ones. 
We evaluate our method on the challenging DocUNet Benchmark dataset, where it sets a series of state-of-the-art records, demonstrating the superiority of our approach compared to existing learning-based solutions. Additionally, we conduct a comprehensive series of ablation experiments to further validate the effectiveness and merits of our method.
\end{abstract}


\begin{figure}[t]
  \centering
  \includegraphics[width=0.97\linewidth]{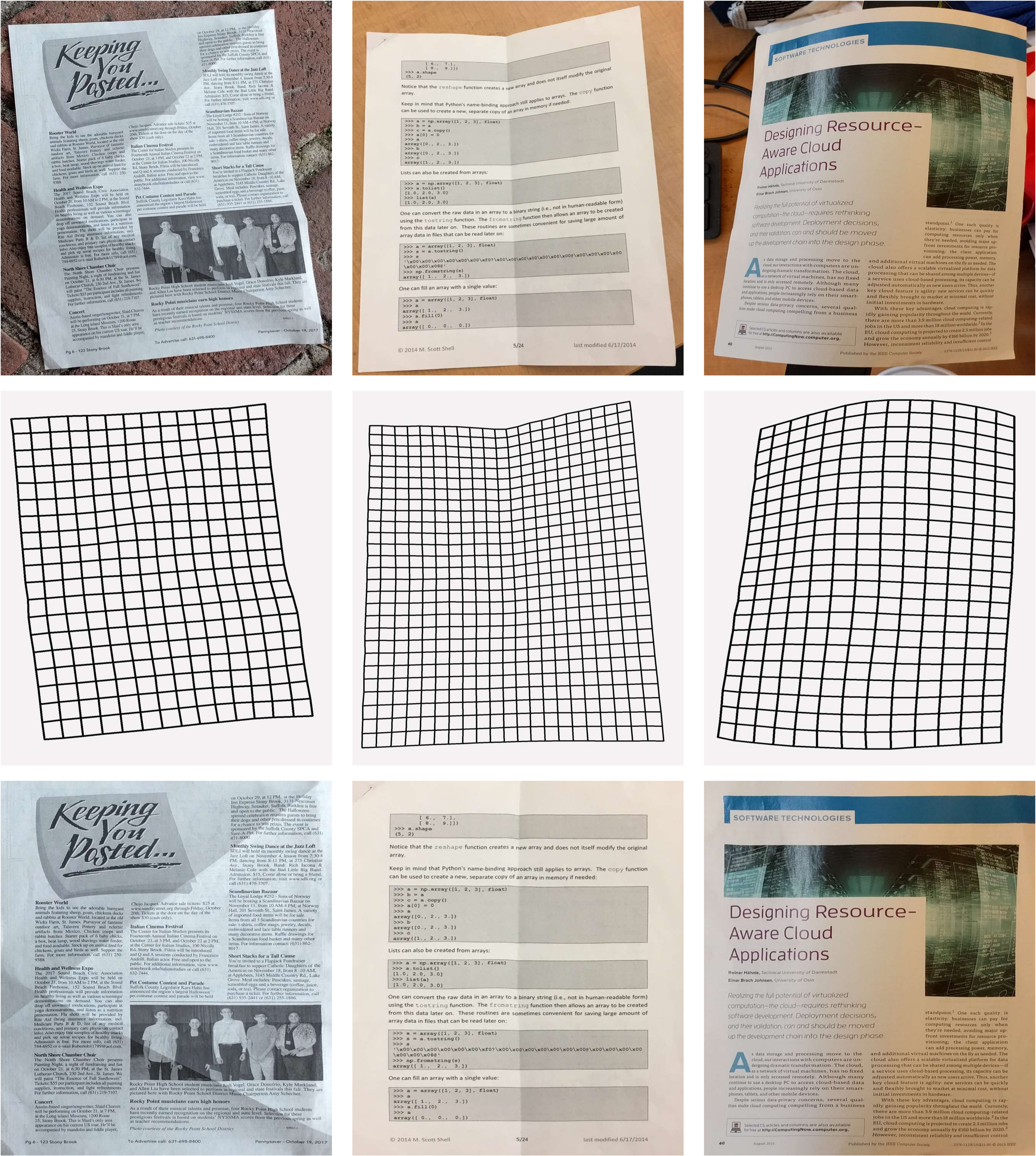}
  \caption{An illustration of rectified results of our DRNet.
  The three rows show the distorted images due to various folds and creases,
  the predicted warping mesh, and our rectified images, respectively.
  }
  \label{fig:overview}

\end{figure}

\section{Introduction}
    \IEEEPARstart{R}ecent years have witnessed, the widespread utilization of smartphones, which engenders an increasing trend of employing them for the digitization of document files. In contrast to traditional flatbed scanners, smartphones provide a more versatile, portable, and uncomplicated option for digitalizing document images. However, the images captured by smartphones are prone to distortion due to various uncontrollable factors, such as physical deformations of the documents, lighting conditions, and camera angles. Such distortions hinder their digital preservation and may adversely affect downstream applications such as automated text recognition~\cite{li2022emu,ciardiello1988experimental,wang2011end,lat2018enhancing,peng2022recognition}, analysis~\cite{yuan2019interpreting,zhang2022multimodal,kim2022ocr}, retrieval~\cite{yu2008semantic,salton1991developments,yang2008harmonizing,liu2023end}, and question answering~\cite{mathew2021docvqa,nie2012beyond,liu2023visual}.

To this end, research on document image rectification, defined as the process of correcting geometric distortions in document images arising from various acquisition processes, has been actively pursued in recent decades.

In academic research, early solutions predominantly employ 3D reconstruction methods. Specifically, these approaches generate a 3D representation of the photographed deformed document, subsequently transforming it into a distortion-free planar surface. To facilitate 3D reconstruction, a subset of methodologies utilizes auxiliary equipment, such as structured-light systems~\cite{brown2004image,937649,6909892} or laser range scanners~\cite{4407722}. Conversely, alternative techniques~\cite{4916075,yamashita2004shape,tsoi2007multi,tsoi2004geometric,Shaodi} obtain multi-perspective images of documents.
Nonetheless, the integration of additional hardware and the requirement for multi-view images inherently constrain their practical implementation. To address these impediments, diverse methods postulate a parametric model for the distorted document surfaces and establish the model through the extraction of particular attributes, encompassing shading~\cite{1561180,wada1997shape}, boundaries~\cite{6628653}, textual lines~\cite{1227630,958227,wu2002document,meng2011metric,luo2022geometric}, or texture flow~\cite{liang2008geometric}. However, the excessively simplified parametric models frequently culminate in a subpar performance, while the optimization procedure incurs a significant computational expenditure.

Recently, deep learning based approaches~\cite{das2020intrinsic,9010747,das2021end,feng2021doctr,jiang2022revisiting,liu2020geometric,8578592,xie2021document,xie2020dewarping,xue2022fourier,feng2021docscanner,feng2022geometric,markovitz2020can,ma2022learning,zhang2022marior,li2019document,verhoeven2023neural,liudocmae2023,feng2023deep} have emerged as potential substitutes for conventional techniques, demonstrating considerable potential in terms of performance and efficiency. Through the estimation of a pixel-wise displacement field~\cite{teed2020raft} that relates the distorted image to its undistorted version using a deep neural network, the distorted document image can be expeditiously unwrapped with the anticipated flow field.
Typically,
DewarpNet \cite{9010747} utilizes 3D shape modeling to improve document unwarping.
DocTr~\cite{feng2021doctr} leverages the capability of transformer~\cite{Vaswani2017AttentionIA} to handle long-distance dependencies, thereby extracting structural or distortion representations of a deformed document.
DocProj~\cite{li2019document} and PWUNet~\cite{das2021end} propose the rectification of local document image patches to achieve finer-grained correction.
DocScanner~\cite{feng2021docscanner} and Marior~\cite{zhang2022marior} iteratively perform multiple corrections to achieve high-precision rectification results.
DocMAE~\cite{liudocmae2023} leverages self-supervised learning with a masked autoencoder to encode structural cues within document images.

However, the intrinsic consistency of document images during the rectification process is overlooked by the above methods.
Existing methods~\cite{feng2021doctr,feng2022geometric,8578592,9010747,liudocmae2023} in the field predominantly utilize warping mesh as their supervisory signal. However, solely computing the difference between the predicted warping mesh and the ground truth one makes the network focus on the learning of the pixel-to-pixel correspondences, thus overlooking the alignment with the distinctive patterns of the undistorted document images and resulting in suboptimal performance. For instance, in an undistorted document image, textlines are almost always horizontally or vertically distributed.
The exploration of a novel supervision manner is also a meaningful research topic.

To address the above issues, we introduce our DRNet, a new deep network for document image rectification. Our DRNet shows three elegant designs.

Firstly, we exploit the intrinsic geometric consistency inherent
in document images to constrain the learning process of distortion
rectification.
Concretely, no matter from which perspective one observes and then corrects the same deformed document, the final appearance will be the same.
Based on this observation,
we constrain the predicted warping mesh when rectifying the different perspectives of the same distorted document image.
Secondly, we propose a deformation refinement module in a coarse-to-fine manner to produce a more accurate rectification result.
Thirdly, we introduce a new perspective for supervising the learning of document rectification networks, where undistorted document images are used for supervision. 
Previous methods~\cite{das2020intrinsic,9010747,das2021end,feng2021doctr,jiang2022revisiting,liu2020geometric,8578592,xie2021document,xie2020dewarping,xue2022fourier,feng2021docscanner,feng2022geometric,markovitz2020can,ma2022learning,zhang2022marior,li2019document,verhoeven2023neural,liudocmae2023,feng2023deep} that rely on warping flow as ground truth may fall short as they primarily focus on geometric alignment, often overlooking semantic consistency. In contrast, by performing pixel-level and semantic-level alignment during training, we ensure both geometric accuracy and semantic consistency, improving the overall rectification outcome.

Extensive experiments are conducted on the challenging DocUNet Benchmark dataset~\cite{8578592}.
The quantitative and qualitative results indicate that our DRNet sets a series of state-of-the-art records, demonstrating the superiority of our method.
In addition, we conduct a substantial number of ablation experiments to demonstrate the effectiveness of our approach.

In summary, we make three-fold contributions as follows:
\begin{itemize}
    \item 
    We present DRNet, 
    a novel deep network for progressive document image rectification that incorporates a geometric self-consistency prior.
    \item 

    We introduce a novel supervised approach for document image restoration where the supervisory signal originates from the undistorted image, rather than the warping mesh.
    \item
    We conduct extensive experiments to validate the merits of DRNet 
    and show state-of-the-art results on the prevalent and proposed benchmarks.
\end{itemize}

\section{Related Work}

We categorize existing research on document image rectification into two principal divisions: (a) rectification based on 3D reconstruction and (b) rectification rooted in deep learning, which will be expounded upon in the following discussion.

\subsection{Rectification Based on 3D Reconstruction}
Traditionally, document image rectification methodologies have relied on 3D reconstruction techniques. Such methods \cite{brown2004image,937649,6909892,4407722} generally encompass the reconstruction of a distorted document page's 3D structure, followed by its conversion into a planar form. For instance, Brown and Seales \cite{937649} employed a structured light 3D acquisition system for reconstructing the 3D model. Zhang et al. \cite{4407722} utilized a laser range scanner and implemented restoration via a physical modeling approach. Meng et al. \cite{6909892} used two structured beams to project onto the deformed document page and obtained its developable surface for rectification purposes. Although these methods have proven effective in real-world contexts, they mandate supplementary hardware for scanning distorted documents, thus limiting their suitability for personal use.

Various approaches exploit multiview images of a document to perform 3D reconstruction. For instance, Tsoi et al. \cite{tsoi2007multi} converted multiview images into a canonical coordinate frame, relying on document boundary interpolation \cite{tsoi2004geometric}. Koo et al. \cite{4916075} employed SIFT \cite{lowe2004distinctive} to match corresponding points between two views, thereby estimating the document surface model. You et al. \cite{Shaodi} presented general developable surface models, capable of depicting a diverse range of paper deformations, built upon ridge-aware 3D reconstruction techniques. Nevertheless, the complexity of image acquisition processes in these works remains a barrier to wider applications.

In order to overcome the limitations mentioned above, other methods focus on reconstructing the 3D model from a single-view image. These approaches initially assume a parametric model for the document surface and proceed to fit the model by extracting specific features, such as shading \cite{1561180,wada1997shape}, boundaries \cite{6628653}, textlines \cite{1227630,958227,wu2002document,meng2011metric,luo2022geometric}, or texture flow \cite{liang2008geometric}. Wada et al. \cite{wada1997shape}, for example, utilized a proximal light source to recover the 3D shape of an unfolded book surface based on shading distribution. He et al. \cite{6628653} derived a book boundary model for the reconstruction of the 3D book surface. Textline-based techniques \cite{1227630,958227,wu2002document,meng2011metric} segmented textlines within distorted document pages to approximate surface model parameters. Additionally, Meng et al. \cite{meng2018exploiting} reconstructed the 3D page shape by leveraging the image's intrinsic vector fields. However, these methods exhibit inherent drawbacks. Predominantly, the fixed parametric models often fall short in tackling the intricate real geometric distortions of deformed documents. Moreover, the optimization procedure incurs a significant computational burden.

\subsection{Deep Learning-Driven Rectification}
Deep learning has recently surfaced as a compelling alternative to traditional techniques. By predicting a pixel-wise displacement field~\cite{teed2020raft} through a deep network, distorted document images can be rectified via resampling pixels from the input image to the adjusted one.

In a groundbreaking contribution, DocUNet~\cite{8578592} utilizes a stacked UNet~\cite{ronneberger2015u} to regress a pixel-wise displacement field, effectively unwarping distorted document images. DocProj~\cite{li2019document} subdivides the input image into several patches, calculates per-patch displacement fields, and subsequently combines them for complete image rectification. Employing a fully convolutional network~\cite{long2015fully}, Xie et al.\cite{xie2020dewarping} introduce an explicit regularization to control the smoothness of the displacement field. Inspired by traditional 3D reconstruction-based approaches, Amir et al.\cite{markovitz2020can} and Das et al.\cite{9010747} incorporate the 3D shape of a distorted document into the network. DocTr\cite{feng2021doctr} adopts transformers~\cite{Vaswani2017AttentionIA} from natural language processing tasks to augment the representation learning of document images.
In contrast to pixel-wise flow regression, DDCP~\cite{xie2021document} predicts only a limited number of control and reference points for rectification, based on the TPS interpolation algorithm~\cite{meijering2002chronology}. 
PWUNet~\cite{das2021end} accounts for the diverse distortion degrees of local patches and estimates local deformation fields for enhanced global unwarping. 

\begin{figure*}[t]
  \centering
  \includegraphics[width=1\linewidth]{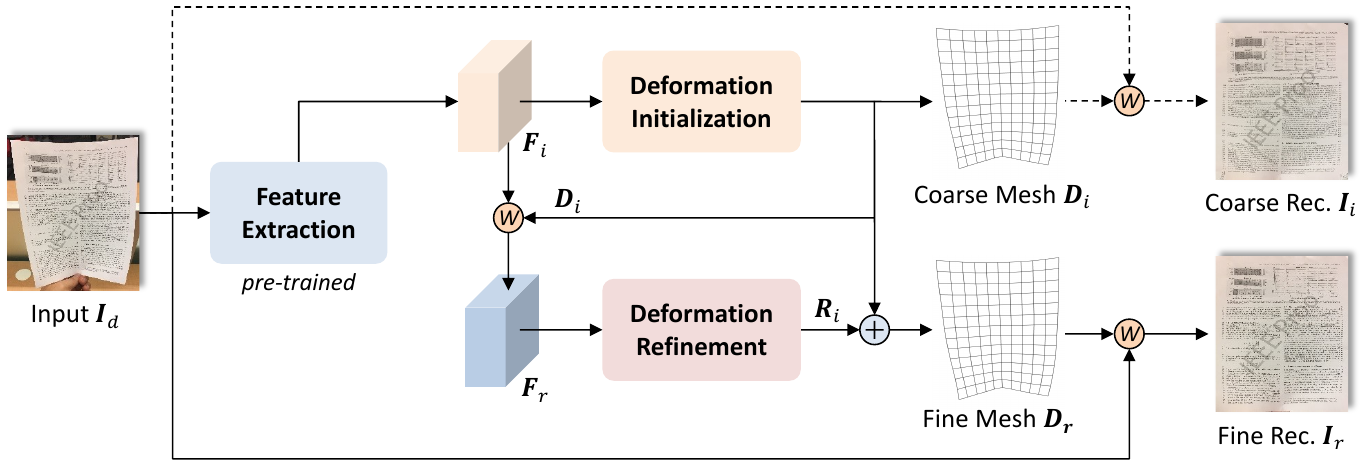}
  \caption{An overview of our proposed DRNet. Given a distorted document image $\bm{I}_d$, DRNet first extracts its features with a pre-trained transformer-based backbone network. Thereafter, the obtained features are fed into a deformation estimation network that performs distortion rectification via a coarse-to-fine manner. Besides, in DRNet, distortion-free document images are used for supervision without the need for warping mesh as ground truth. The dashed line signifies a process that is exclusive to the training phase, while the solid line indicates a process that occurs during the inference phase.}
  \label{fig:pipeline}
\end{figure*}

DocGeoNet~\cite{feng2022geometric} suggests encoding representations of rectification cues to refine rectification by detecting text lines within the network. RDGR~\cite{jiang2022revisiting} firstly identifies text lines and document boundaries, and subsequently acquires the rectified image by solving an optimization problem grounded in the proposed grid regularization. FDRNet~\cite{xue2022fourier} centers on high-frequency components in Fourier space to extract structural information from deformed documents for improved rectification. Marior~\cite{zhang2022marior} tackles document images with large marginal areas and incrementally rectifies them to achieve a stable state. DocScanner~\cite{feng2021docscanner} strives to rectify distortion iteratively and progressively utilizing a lightweight network. PaperEdge~\cite{ma2022learning} bolsters document unwarping performance by integrating real-world document images during training. 
DocTr-Plus~\cite{feng2023deep} proposes a generic pipeline for processing various document images in the wild, regardless of the document boundaries completeness in the captured image.
CGU-Net~\cite{verhoeven2023neural} introduces a novel grid-based method for document unwarping and presents a new synthetic dataset called UVDoc. 
IllTrTemplate~\cite{hertlein2023template} addresses the challenge of illumination correction for document images by integrating prior visual cues from document templates, especially considering the complexities arising from imperfectly unwarped documents. FTDR~\cite{li2023foreground} emphasizes the foreground and text-line regions, and employs a global-local fusion approach with cross attention to enhance the document's readability and rectification effect.

Distinct from prior deep learning solutions, our DRNet leverages geometric consistency and a unique supervisory perspective, employing a coarse-to-fine approach to deliver a more efficient and robust rectification of document images.

\section{Method}
\subsection{Architecture Overview}
Fig.~\ref{fig:pipeline} provides an overview of our DRNet architecture.
Given a distorted document image $\bm{I}_d$, 
we first extract its feature via a Vision Transformer (ViT)~\cite{ViT}.
It is pre-trained with the large-scale document image dataset~\cite{liudocmae2023} via a self-supervised manner~\cite{he2022masked} to model fine-grained distortion/structure features of a deformed document.
Thereafter, we feed the extracted features to the subsequent deformation prediction network that performs the rectification via a coarse-to-fine manner.
Different from existing methods~\cite{8578592,9010747,feng2021doctr,feng2022geometric,feng2021docscanner,liudocmae2023} that apply supervision to the predicted warping mesh,
we propose a novel perspective, asserting that the undistorted image can also serve as a supervisory signal.
Technically, we directly supervise the output of the network with an undistorted image, forcing the network to learn pixel-wise mapping to rectify the distorted input.
In the following sections, we detail our proposed DRNet.

\subsection{Feature Extraction}
Given a distorted document image $\bm{I}_d \in\mathbb{R}^{H\times W\times3}$,
where $H$ and $W$ denote the height and width of this \emph{RGB} image,
we first remove its background with a document localization module~\cite{qin2020u2,feng2021docscanner}.
Then, we feed the obtained image $\bm{I}_e$ into a backbone network for distortion feature extraction.
It is based on the Vision Transformer (ViT)~\cite{ViT} and 
consists of $N_b$ transformer encoder layers~\cite{Vaswani2017AttentionIA}.
Besides, following DocMAE~\cite{liudocmae2023}, we pre-train the backbone network from scratch with the masked image modeling strategy (MIM),
$i.e.$,
masking random patches of the
input distorted document images and reconstruct the missing pixels.
To accomplish such a pretext task,
the network learns to extract distortion/structure features about the input images by the reconstruction of textlines and document boundaries, benefiting the downstream distortion rectification.

Specifically,
in accordance with ViT~\cite{ViT}, we begin by partitioning the document image $\boldsymbol{I}_e \in \mathbb{R}^{H \times W \times 3}$ into a sequence of 2D patches $\boldsymbol{x}_p \in \mathbb{R}^{N_p \times\left(P^2 \cdot 3\right)}$.
Here, $P$ represents the patch size, and the value of $N_p$, which denotes the number of patches and serves as the input sequence length for the
transformer encoder~\cite{ViT}, is computed as $N_p=H W / P^2$.
Following previous methods~\cite{feng2021doctr,liu2020geometric,feng2021docscanner,liudocmae2023},
the input images are stripped of their noisy backgrounds using a U-structure segmentation network~\cite{qin2020u2},
aiming to avoid the extra implicit learning to
identify the foreground document for predicting the rectified document.

The flattened patches $\boldsymbol{x}_p \in \mathbb{R}^{N_p \times\left(P^2 \cdot 3\right)}$ are first mapped to $C_t$ hidden dimensions with a trainable linear projection~\cite{ViT} with added
positional embeddings~\cite{Vaswani2017AttentionIA,he2022masked}.
Subsequently,
it is fed into $N_b$ transformer encoder layers~\cite{Vaswani2017AttentionIA}.
Thereby, we obtain the feature $\boldsymbol{F}_i \in \mathbb{R}^{N_p \times C_t}$, which is fed into the subsequent deformation prediction network.

\subsection{Deformation Initialization}
As shown in Fig.~\ref{fig:pipeline}, given the feature $\boldsymbol{F}_i \in \mathbb{R}^{N_p \times C_t}$,
we first inject it into another $N_i$ transformer encoder layers~\cite{Vaswani2017AttentionIA}
and obtain the feature $\boldsymbol{F}_i' \in \mathbb{R}^{N_p \times C_t}$.
Then,
$\boldsymbol{F}_i'$ is fed into a mesh estimation head based on convex upsampling~\cite{feng2021doctr},
which produces a coarse deformation mesh $\boldsymbol{D}_i \in \mathbb{R}^{H \times W \times 2}$.
Finally, we unwarp the input image with $\boldsymbol{D}_i$ and obtain the coarse rectified image $\boldsymbol{I}_i \in \mathbb{R}^{H \times W \times 3}$.

Concretely, we transform the feature matrix $\bm{F}_i' \in \mathbb{R}^{N_p \times C_t}$ into the shape form ${\frac{H}{P} \times \frac{W}{P} \times 2}$. Subsequently, we employ a learnable upsample module~\cite{feng2021doctr} to elevate the $\frac{1}{P}$ scale warping mesh to its full-scale counterpart $\boldsymbol{D}_i \in \mathbb{R}^{H \times W \times 2}$. It is imperative to note that $\boldsymbol{D}_i$ represents a dense flow field delineating the pixel-wise coordinate displacement from the distorted image to the distortion-free counterpart.
Upon obtaining the predicted warping mesh $\boldsymbol{D}_i \in \mathbb{R}^{H \times W \times 2}$, we proceed to resample the pixels from the background-excluded $\boldsymbol{I}_{e}$, thereby generating the rectified one $\boldsymbol{I}_{i} \in \mathbb{R}^{H \times W \times 3}$, as demonstrated below,
\begin{equation}\label{equ:task}
\bm{I}_i(x,y) = \bm{I}_e(\bm{D}^u_i(x,y), \bm{D}^v_i(x,y)),
\end{equation}
where $(x,y)$ denotes the integer pixel coordinates within the rectified image $\bm{I}_i$; $\bm{D}^u_i$ and $\bm{D}^v_i$ signify the coordinate mapping in the horizontal and vertical directions, respectively; and $(\bm{D}^u_i(x,y), \bm{D}^v_i(x,y))$ corresponds to the projected pixel coordinate within the distorted image $\bm{I}_e$.

\subsection{Deformation Refinement}
In this module, we further refine the estimated deformation mesh $\boldsymbol{D}_i \in \mathbb{R}^{H \times W \times 2}$.
Given the feature $\bm{F}_i \in \mathbb{R}^{\frac{H}{P} \times \frac{W}{P} \times 2}$,
we first transform it to the rectification domain based on mesh $\boldsymbol{D}_i$.
Importantly, similar to the image warping manner in Eq.~\eqref{equ:task}, this transformation operation is also
implemented based on bilinear interpolation. Therefore,
the gradients can be computed to the input feature $\bm{F}_i$ and warping mesh $\boldsymbol{D}_i$ for backpropagation, based on the classic STN~\cite{jaderberg2015spatial}.

The transformed feature $\bm{F}_r \in \mathbb{R}^{\frac{H}{P} \times \frac{W}{P} \times 2}$ is further fed into another $N_r$ transformer encoder layers~\cite{Vaswani2017AttentionIA} and a learnable upsample module~\cite{feng2021doctr}, which produces a full-scale residual map $\bm{R}_i \in \mathbb{R}^{{H} \times {W} \times 2}$. The upsample manner is the same as that in the previous deformation initialization stage.
The residual map $\bm{R}_i$ is utilized to update the mesh $\bm{D}_i$ as follows,
\begin{equation}\label{equ:resi}
\bm{D}_r = \bm{D}_i + \bm{R}_i,
\end{equation}
where $\boldsymbol{D}_r$ denotes the refined deformation mesh. 
With $\boldsymbol{D}_r$, we unwarp the distorted input $\boldsymbol{I}_e$ based on Eq.~\eqref{equ:task}, and produce the final rectified document image $\boldsymbol{I}_r$.

\begin{figure}[t]
  \centering
  \includegraphics[width=1\linewidth]{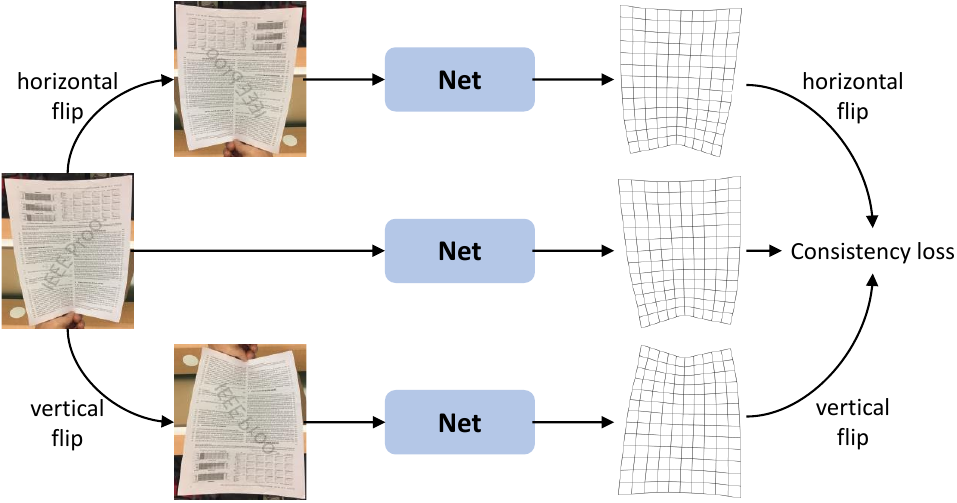}
  \caption{An illustration of the proposed consistency loss to facilitate document image rectification.}
  \label{fig:consistency}
 
\end{figure}

\subsection{Training Objective}~\label{loss}
During the training stage, the whole network is end-to-end optimized with the following objective,
\begin{equation}
	\mathcal{L} = \mathcal{L}_i +  \mathcal{L}_r,  
\end{equation}
where $\mathcal{L}_i$ and $\mathcal{L}_r$ denote the loss on the deformation initialization and refinement stage, respectively.
Both $\mathcal{L}_i$ and $\mathcal{L}_r$ consist of two terms. Take $\mathcal{L}_i$ as the example: 
\begin{equation}\label{lclb}
	\mathcal{L}_i = \mathcal{L}_c + \mathcal{L}_w, 
\end{equation}
where $\mathcal{L}_c$ and $\mathcal{L}_w$ represent the self-consistency loss and the image rectification loss, respectively.
We detail them next. 

\smallskip
1) Self-consistency loss $\mathcal{L}_c$:
Considering that 
no matter from which perspective one observes and then corrects the same wrinkled document, the final rectified appearance will be the same.
Thereby,
we propose a self-consistency loss.
Specifically,
as shown in Fig.~\ref{fig:consistency},
during the training phase, we first randomly flip the input image $\bm{I}_d$ either horizontally or vertically (termed as $T(*)$), and subsequently feed both the original and flipped images into the network.
Then, 
based on their geometric consistency,
we further calculate the following loss on the predicted warping mesh pair:
\begin{equation}
	\mathcal{L}_c = \left \| \bm{D}_{i}^{(o)} - T(\bm{D}_{i}^{(f)}) \right \|_1,
\end{equation}
where $\bm{D}_{i}^{(o)}$ and $\bm{D}_{i}^{(f)}$ denote the output warping mesh of the original and flipped distorted image, respectively.
Such a constraint makes the network produce a more accurate and robust rectification result.
In this way, our supervision is free of explicit warping flow as ground truth.
Note that different from the flip operation for the input image $\bm{I}_d$,
the flip of a mesh involves another pointed direction reversal in the corresponding channel.

\smallskip
2) Image rectification loss $\mathcal{L}_w$: 
Different from existing methods~\cite{8578592,9010747,feng2021doctr,feng2022geometric,feng2021docscanner,liudocmae2023} that apply supervision to the predicted warping mesh, we argue that the warping mesh is not a necessary supervision signal for the task and introduce a new supervision manner.
It is defined based on the perspectives of low-level pixel alignment and high-level semantic alignment.
Concretely, our DRNet calculates the loss on the rectified images as follows,

\begin{equation}\label{eq:lw}
\mathcal{L}_{w} = \mathcal{L}_p + \mathcal{L}_v = \left \| \bm{I}_{gt} - \bm{I}_i \right \|_1 + \left \| V(\bm{I}_{gt}) - V(\bm{I}_i) \right \|_1,
\end{equation}
where the first term $\mathcal{L}_p$ denotes the $\mathcal{L}_{1}$ loss between the predicted rectified image $\bm{I}_i$ and its given ground truth $\bm{I}_{gt}$, and the second term is the VGG loss $\mathcal{L}_v$ that is also known as perceptual loss or content loss. Here $V(\cdot)$ denotes a  pre-trained 19-layer VGG network~\cite{simonyan2015deep,zhou2022hrinversion}. The VGG loss is defined as the $L_1$ distance between the output feature map of the ReLU activation layers for estimated rectified image $\bm{I}_i$ and ground truth image $\bm{I}_{gt}$.
Such a constraint compels the network to output a rectified image that is both locally and globally consistent with the undistorted one.

\section{Datasets and Metrics}
In this section,
we provide a detailed description of the used datasets and evaluation metrics below.

\subsection{Datasets}
Our DRNet is pre-trained using the LDIR dataset~\cite{liudocmae2023} for representation learning.
Then, we train the network with the Doc3D dataset~\cite{9010747} for the learning of document image rectification.
The evaluation is performed on the challenging DocUNet Benchmark dataset~\cite{8578592}.

\smallskip
\textbf{LDIR.}
The LDIR dataset~\cite{liudocmae2023} is a comprehensive synthetic dataset designed for document image rectification through self-supervised learning. By employing 3D rendering software\footnote{https://www.blender.org/}, it accurately emulates real-world document textures, illumination conditions, backgrounds, and distortions, thus ensuring the superior quality of LDIR. This dataset comprises 200,000 distorted document images, which are derived from authentic document data commonly encountered in daily life, such as various books and magazines.

\smallskip
\textbf{Doc3D.} The Doc3D dataset~\cite{9010747} is generated using real document data and rendering software. The dataset encompasses 100,000 distorted document images, each accompanied by ground truth data such as a 3D coordinate map, albedo map, normals map, depth map, forward warping flow map, and backward warping flow map.

\smallskip
\textbf{DocUNet Benchmark.} The rigorous DocUNet Benchmark dataset~\cite{8578592} is a prevalent evaluation benchmark for document image rectification. It contains 130 photographs of authentic documents captured by mobile cameras. The dataset features a diverse array of document types, including receipts, letters, fliers, magazines, academic papers, books, and more. Moreover, the documents display varying degrees of distortion and background complexity, covering a range of difficulty levels.

It is important to note that in the DocUNet Benchmark dataset~\cite{8578592} the $127^{th}$ and $128^{th}$ distorted document images are rotated by 180 degrees, leading to a mismatch with the ground truth documents. This inconsistency has been overlooked by prior methods~\cite{9010747,das2021end,feng2021doctr,jiang2022revisiting,liu2020geometric,8578592,xie2021document,xie2020dewarping,xue2022fourier,markovitz2020can,ma2022learning,zhang2022marior,li2019document}. In the present study, the reported performance metrics of our DRNet and other methods are based on the corrected dataset.

\setlength{\tabcolsep}{1.3mm}
\begin{table}[t]
\centering
	\caption{Quantitative comparisons with existing methods on the DocUNet Benchmark dataset~\cite{8578592}. ``$\uparrow$'' indicates the higher the better and ``$\downarrow$'' means the opposite. The best and second-best results are highlighted in \textbf{bold} and \underline{underlined}, respectively.}
\begin{tabular}{l|c|cccc} 
   \toprule
   \textbf{Methods} & \textbf{Venue} &\textbf{MSSIM} $\uparrow$ &\textbf{LD} $\downarrow$  &\textbf{ED} $\downarrow$ &\textbf{CER} $\downarrow$   \\ 
   \midrule
    Distorted & / & 0.25 & 20.51 & 2111.56 & 0.5352 \\
    \midrule
    DocUNet~\cite{8578592} & \emph{CVPR'18} & 0.41 & 14.19 & 1933.66 & 0.4632 \\
    DocProj~\cite{li2019document} & \emph{TOG'19} & 0.29 & 18.01  & 1712.48 & 0.4267 \\
    FCN-based~\cite{xie2020dewarping} & \emph{DAS'20}  & 0.45 & 7.84 & 1792.60 & 0.4213 \\ 
    DewarpNet~\cite{9010747} & \emph{ICCV'19}  & 0.47 & 8.39  & 885.90 & 0.2373 \\
    DocTr~\cite{feng2021doctr} & \emph{MM'21} &   \underline{0.51} & 7.76  & 724.84 & 0.1832 \\
    PWUNet~\cite{das2021end} & \emph{ICCV'21} & 0.49 & 8.64  & 1069.28 & 0.2677 \\    
    DDCP~\cite{xie2021document}  & \emph{ICDAR'21} & 0.47 & 8.99 & 1442.84 & 0.3633 \\
    DocGeoNet~\cite{feng2022geometric} & \emph{ECCV'22} &  0.50  & 7.71 &  \underline{713.94}  &  0.1821  \\ 
    Marior~\cite{zhang2022marior} & \emph{MM'22} & 0.48 &   \underline{7.44}  & 776.22 & 0.1928 \\
    RDGR~\cite{jiang2022revisiting} & \emph{CVPR'22} &  0.50  & 8.51 & 729.52  &  \underline{0.1717}   \\
    FDRNet~\cite{xue2022fourier}  &  \emph{CVPR'22} &  \textbf{0.54}  & 8.21 & 829.78 & 0.2068     \\ 
    PaperEdge~\cite{ma2022learning} & \emph{SIGGRAPH'22} & 0.47 & 7.99 &  777.76  &  0.2014  \\
    DocMAE~\cite{liudocmae2023} & \emph{ICME'23}   &  0.50  &  7.63  & 801.52 & 0.2019\\
    \midrule
    DRNet & - & \underline{0.51} & \textbf{7.42} & \textbf{644.48} & \textbf{0.1649}  \\  
   \bottomrule
\end{tabular}
\label{tab:t1}
\end{table} 

\subsection{Metrics}
We evaluate our method based on the classic metrics in the field, including MSSIM, LD, ED, and CER. They evaluate the rectified images in terms of image quality, distortion rectification, and OCR accuracy.

\smallskip
\textbf{MSSIM.}
The Structural SIMilarity (SSIM) index~\cite{1284395} assesses the per-patch similarity between two images. As the perceptibility of details depends on the image's sampling density, the Multi-scale Structural SIMilarity (MSSIM)~\cite{1292216} is proposed and calculates a weighted summation of SSIM~\cite{1284395} across multiple scales. In the context of document image rectification, both rectified and ground truth flatbed-scanned images are initially resized to an area of 598,400 pixels, in accordance with the recommendation of DocUNet~\cite{8578592}. Subsequently, MSSIM is computed using a 5-level pyramid, adhering to the original MSSIM methodology~\cite{1292216}.

\setlength{\tabcolsep}{4mm}
\begin{table*}[t]
	
	\caption{Ablation experiments of the training strategy of DRNet in terms of image similarity, distortion metrics, and OCR performance on the DocUNet Benchmark dataset~\cite{8578592}. ``$\uparrow$'' indicates the higher the better and ``$\downarrow$'' means the opposite.}
	
	\centering
	\begin{tabular}{c|c|c|c|c|c|cc}  
        \toprule
		\multicolumn{4}{c|}{Method}  & \multirow{2}*{\textbf{MSSIM} $\uparrow$} & \multirow{2}*{\textbf{LD} $\downarrow$} & \multirow{2}*{\textbf{ED} $\downarrow$} & \multirow{2}*{\textbf{CER} $\downarrow$} \\
        \cmidrule{1-4}
        & Baseline  & + Self-consistency loss $\mathcal{L}_c$&  + Refinement &  &  &  &   \\  
        \midrule
	(a) & \scriptsize{\Checkmark} & & & 0.50 & 7.63 & 801.52 & 0.2019 \\  
        \midrule 
        (b) & \scriptsize{\Checkmark} & \scriptsize{\Checkmark} & & \textbf{0.52} &  7.57   & 722.14  &  0.1808 \\ 
        \midrule 
        (c) & \scriptsize{\Checkmark} & \scriptsize{\Checkmark} & \scriptsize{\Checkmark} &  0.51   &  \textbf{7.42} & \textbf{644.48} & \textbf{0.1649}  \\         \bottomrule		
	\end{tabular}
	
	\label{aba:aba}
\end{table*}

\setlength{\tabcolsep}{4mm}
\begin{table*}[t]
	
	\caption{Ablation experiments of supervision setting of DRNet in terms of image similarity, distortion metrics, and OCR performance on the DocUNet Benchmark dataset~\cite{8578592}. ``$\uparrow$'' indicates the higher the better and ``$\downarrow$'' means the opposite.}
	
	\centering
	\begin{tabular}{c|c|c|c|c|c|cc}  
        \toprule
		\multicolumn{4}{c|}{Method}  & \multirow{2}*{\textbf{MSSIM} $\uparrow$} & \multirow{2}*{\textbf{LD} $\downarrow$} & \multirow{2}*{\textbf{ED} $\downarrow$} & \multirow{2}*{\textbf{CER} $\downarrow$} \\
        \cmidrule{1-4}
          & loss on warping mesh   & pixel-wise $L_1$ loss $\mathcal{L}_p$ &  VGG loss $\mathcal{L}_v$  &  &  &  &   \\  
        \midrule
	(a) &  \scriptsize{\Checkmark} & & & \textbf{0.51}
 & 7.58
 & 712.12
 & 0.1682
 \\  
        \midrule 
        (b) &  & \scriptsize{\Checkmark} & &   0.49  &  8.81   & 1135.50  &  0.2887 \\ 
        \midrule 
        (c) &  &  & \scriptsize{\Checkmark} &  0.50   &  7.69 & 804.76 & 0.2036 \\      
         \midrule    
        (d) &  & \scriptsize{\Checkmark} & \scriptsize{\Checkmark} & \textbf{0.51} &  \textbf{7.42} & \textbf{644.48} & \textbf{0.1649}  \\    
         \bottomrule		
	\end{tabular}
	
	\label{aba:loss}
\end{table*}

\setlength{\tabcolsep}{3mm}
\begin{table}[t]
	\caption{Ablation experiments of the refinement times in terms of image similarity, distortion metrics, and OCR performance on the DocUNet Benchmark dataset~\cite{8578592}. ``$\uparrow$'' indicates the higher the better and ``$\downarrow$'' means the opposite.}
	\centering
	\begin{tabular}{c|c|c|c|c}
        \toprule
		\textbf{Refinement Times} & \textbf{MSSIM} $\uparrow$ & \textbf{LD} $\downarrow$ & \textbf{ED} $\downarrow$ & \textbf{CER} $\downarrow$ \\
        \midrule
		baseline & \textbf{0.52} &  7.57   & 722.14  &  0.1808 \\  
        \midrule 
        1 &  0.51  &  \textbf{7.42} & \textbf{644.48} & \textbf{0.1649} \\ 
        \midrule 
        2 & \textbf{0.52}  &  7.58 & 715.82 & 0.1820 \\      
        \midrule    
        3 & \textbf{0.52}  &  7.69 & 757.24 & 0.1796  \\    
        \bottomrule		
	\end{tabular}
	\label{abc:iter}
\end{table}

However, MSSIM~\cite{1292216} is not an ideal metric for evaluating geometric rectification tasks~\cite{feng2022geometric}. The MSSIM metric is suitable for image restoration tasks where pixel-level alignment is paramount, such as deraining~\cite{ren2019progressive} and denoising~\cite{buades2005review}. They excel at identifying artifacts, including blurs, noise, color shifts, and halos. However, they might not be adept at capturing geometric distortion, which pertains to the misalignment of pixels between a rectified and an undistorted reference image. DocGeoNet~\cite{feng2022geometric} supports the continued use of this metric in future document rectification research but advises against using it as the primary criterion for performance evaluation.

\smallskip
\textbf{LD.}
Local Distortion (LD)~\cite{Shaodi} initially matches the rectified image with the ground truth image by employing a dense SIFT-flow~\cite{5551153}, denoted by $(\Delta \bm{x}, \Delta \bm{y})$. Here, $\Delta \bm{x}$ and $\Delta \bm{y}$ represent the horizontal and vertical displacement maps of the matched pixels from the ground truth image to the rectified image, respectively. LD is defined as the mean $L_2$ distance among all matched pixels, quantifying the average local deformation of the rectified image.

\smallskip
\textbf{ED and CER.}
Edit Distance (ED)~\cite{levenshtein1966binary} quantifies the dissimilarity between two strings, defined as the minimum number of operations necessary to transform one string into the reference string. This measure can be efficiently calculated using a dynamic programming algorithm. The involved operations encompass deletions $(d)$, insertions $(i)$, and substitutions $(s)$. Subsequently, the Character Error Rate (CER) can be computed as:

\begin{equation}\label{equ:cer}
CER=(d+i+s)/{N_c} ,
\end{equation}

\noindent
where $N_c$ represents the number of characters in the reference string. CER indicates the percentage of characters in the reference text that are incorrectly recognized in the rectified image. A lower CER value signifies the better performance of the rectification method, with 0 being a perfect score.

\begin{figure}[t]
  \centering
  \includegraphics[width=0.98\linewidth]{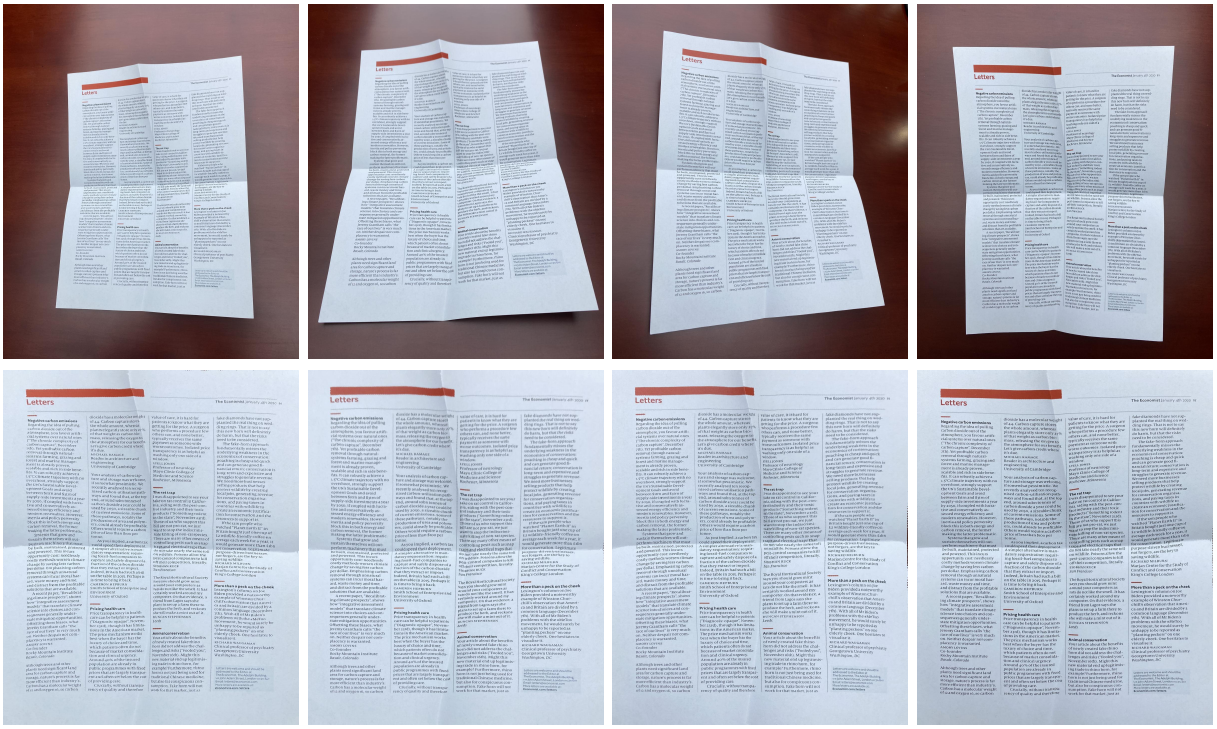}
  \caption{Robustness experiments about the shooting view. The first row shows the input distorted images captured with various shooting views, and the second row presents the rectified images of our DRNet.}
  \label{fig:view}

\end{figure}

Note that the reference images for the document image rectification task are not standard references in the strictest sense. Actually, they are undistorted electronic versions~\cite{8578592}, akin to PDF documents on computers, while our estimated restored images come with noise such as illumination and shadows introduced by the camera capture process. Therefore, existing metrics like MSSIM, LD, and OCR accuracy might not fully reflect the performance of a rectification method. We argue that an important research direction in this field is to explore reference-free evaluation metrics. One likely idea is to detect all horizontal textlines in the rectified image. If a textline is perfectly horizontal, it signifies a perfect restoration. Otherwise, the overall tilt of the textlines can measure the quality of rectification.

\section{Experiments}
\subsection{Implementation Detail}

\smallskip
\textbf{Training Details.}
During the training phase, the image size $(H, W)$ is set to $(288, 288)$.
The transformer layer number $N_b$, $N_i$, and $N_r$ are set as 6. 
Besides, the patch size $P$ is 16, and the latent dimension in transformer $C_t$ is 512.

The training process employs the AdamW optimizer~\cite{loshchilov2017decoupled} in conjunction with the OneCycle learning rate schedule~\cite{smith2019super}, utilizing a maximum learning rate of $10^{-4}$. The warm-up phase constitutes $10\%$ of the total epochs. The network is trained for a total of 65 epochs using a batch size of 8. Training is conducted on 2 NVIDIA GeForce RTX 3090 GPUs. Additionally, to enhance generalization to real data characterized by complex illumination conditions, jitter is introduced in the HSV color space, amplifying variations in illumination and document color.

\smallskip
\textbf{Inference Methodology.}
In the evaluation of the inference phase, real-world distorted document images with arbitrary resolutions are initially resized to the dimensions of $(288, 288)$. These processed images are subsequently input into the network, which produces a warping mesh of shape $(288, 288)$. The warping mesh is then resized to match the original dimensions of the input distorted image based on bilinear interpolation, and the distorted image is then unwarped using the derived warping mesh. Through this process, a high-resolution, rectified document image with legible text is produced as the output.

\begin{figure}[t]
  \centering
  \includegraphics[width=0.98\linewidth]{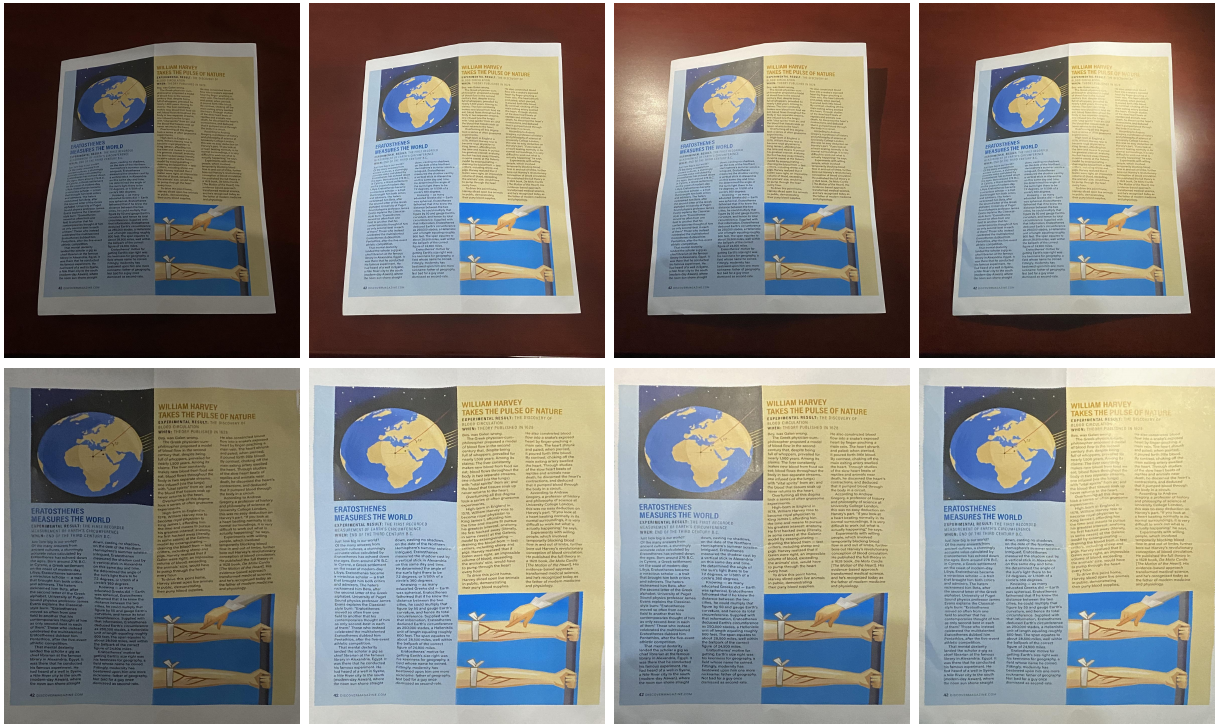}
  \caption{Robustness experiments about the illumination. The first row shows the input distorted images captured under different illumination conditions, and the second row presents the rectified images of our DRNet.}
  \label{fig:illumination}

\end{figure}  

\begin{figure*}[t]
  \centering
  \includegraphics[width=0.98\linewidth]{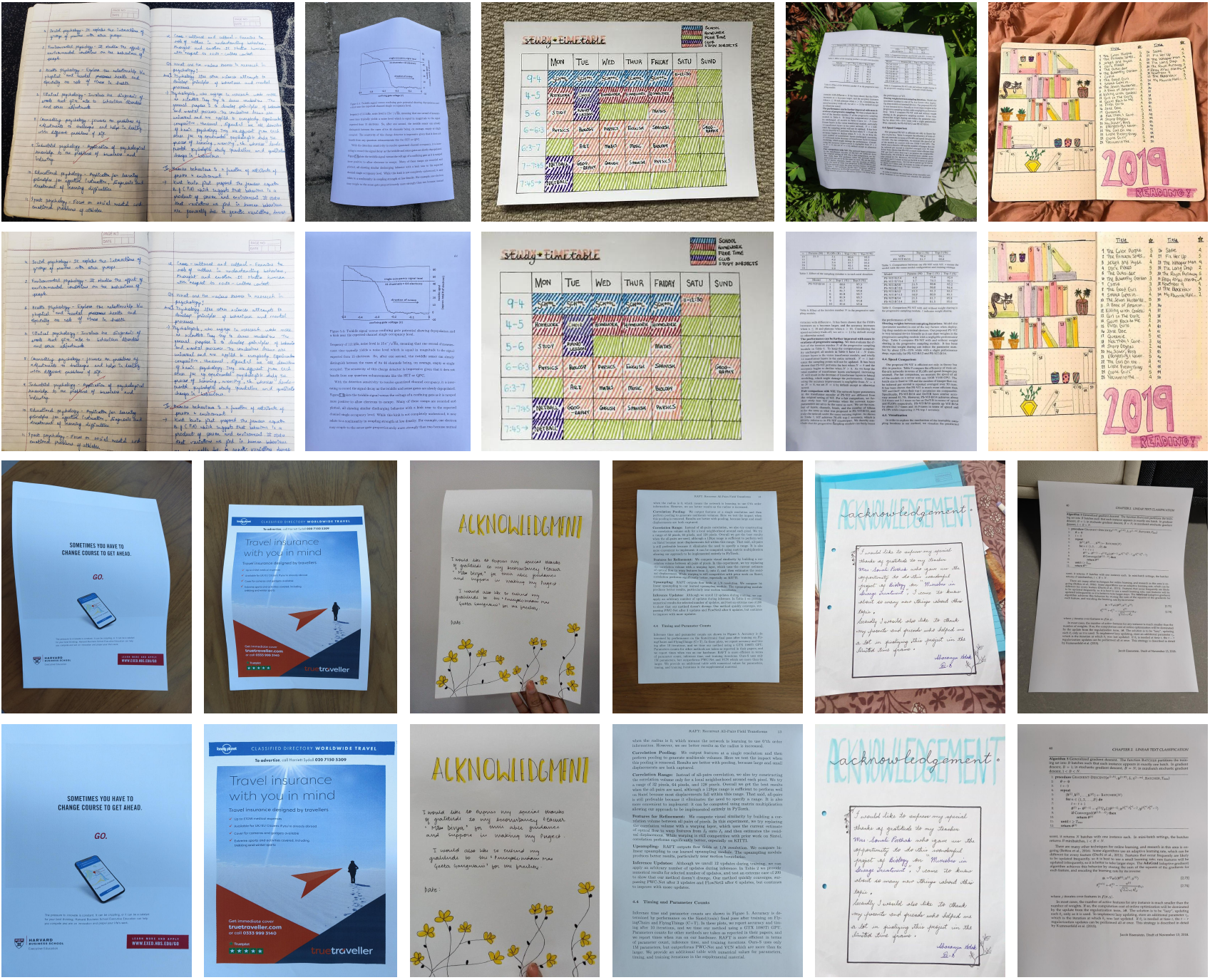}
  \caption{Qualitative results of the proposed DRNet on various real-world distorted document images. Both printed and handwritten cases are included. The first and third rows present the distorted document images, and the second and fourth rows illustrate the corresponding rectified results.}
  \label{fig:showcase}

\end{figure*} 

\begin{figure*}[t]
  \centering
  \includegraphics[width=1\linewidth]{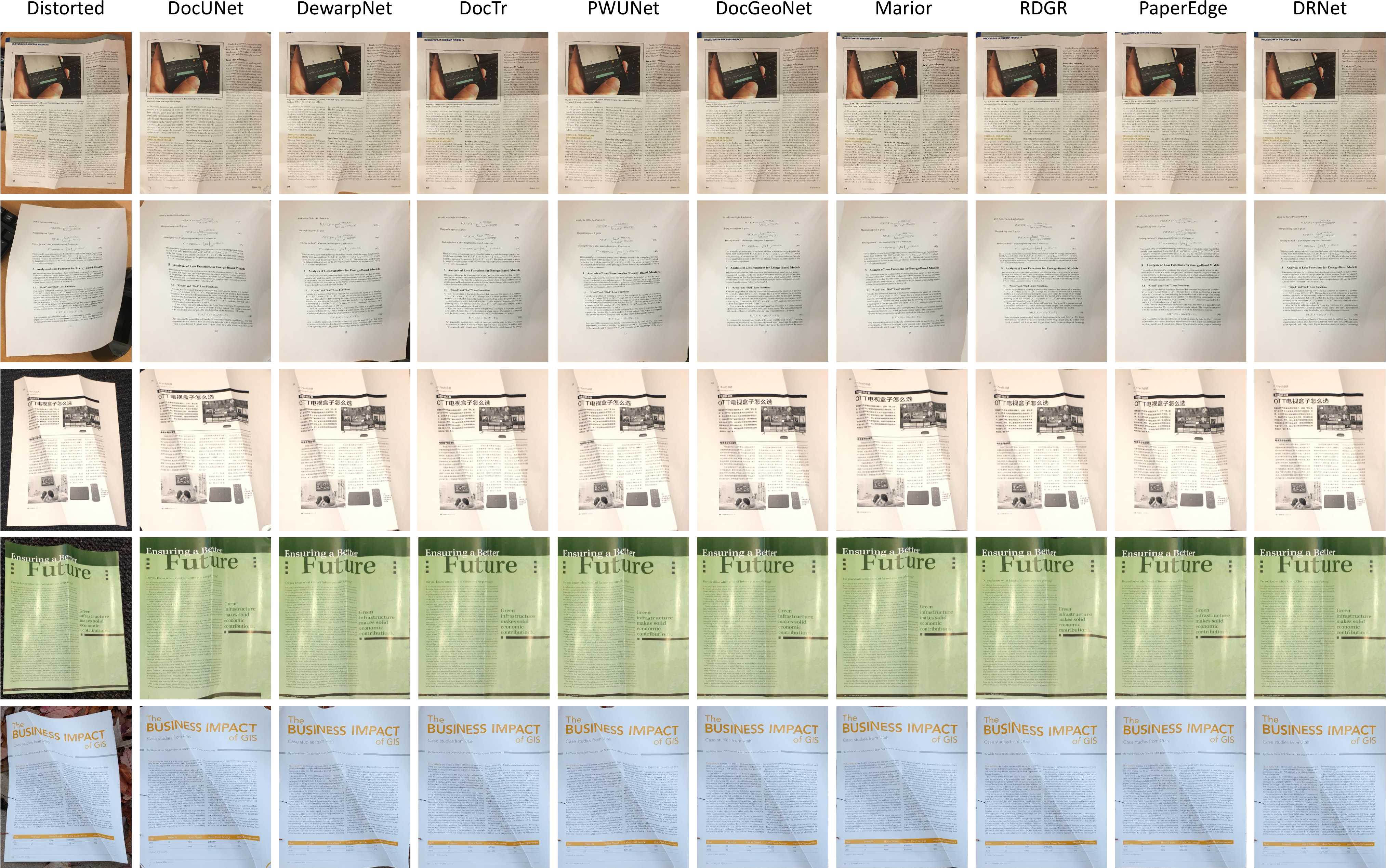}
  \caption{Qualitative comparisons with existing learning-based methods, including original distorted images, DocUNet~\cite{8578592}, DewarpNet~\cite{9010747}, DocTr~\cite{feng2021doctr}, PWUNet~\cite{das2021end}, DocGeoNet~\cite{feng2022geometric}, Marior~\cite{zhang2022marior}, RDGR~\cite{jiang2022revisiting}, PaperEdge~\cite{ma2022learning}, and our DRNet, from left to right.}
  \label{fig:viscom}
  
\end{figure*} 

\smallskip
\textbf{Evaluation Procedure.}
To compute the MSSIM~\cite{1292216}, the weights for each level of the image pyramid are set as follows: 0.0448, 0.2856, 0.3001, 0.2363, and 0.1333, in accordance with the methods described in~\cite{feng2021doctr,feng2021docscanner,feng2022geometric,liudocmae2023}. The Matlab version employed is 2019a. Subsequently, Tesseract (v5.0.1)~\cite{4376991} is utilized as the OCR engine to recognize text strings in both the rectified and ground truth document images respectively, following methods~\cite{feng2021doctr,feng2021docscanner,feng2022geometric}. 

For the DocUNet Benchmark dataset~\cite{8578592}, 50 images are selected for the OCR evaluation, following methods~\cite{9010747, feng2022geometric, feng2021docscanner}. The selected images primarily contain text. This is because a low character count ${N_c}$ in Eq.~\eqref{equ:cer} would result in significant variance for the CER when text is scarce within an image.

\subsection{Ablation Study}
We conduct extensive experiments to validate the core components and settings of our method.

\smallskip
\textbf{Impact of Self-consistency Loss.}
We first validate the proposed self-consistency loss.
As mentioned in Sec.~\ref{loss},
during training,
we impose consistency constraints on the predicted meshes of two deformed document images captured from distinct perspectives.
In the ablation study, we exclude the computation of this particular loss.
The results are shown in Table~\ref{aba:aba}.
As we can see,
without the proposed self-consistency loss,
the performance becomes worse.
The experimental results can be attributed to the computation of consistency loss for the same image captured from different perspectives, which enables the network to effectively leverage the visual clues, thus producing more accurate and robust rectification results.

Our consistency loss diverges from traditional data augmentation approaches. Conventional data augmentation aims to enrich the diversity of input data, covering various scenarios to bolster generalization performance on test sets. Take the document image correction task as an example: a commonly employed data augmentation strategy is color jitter, which exposes the document image to an array of lighting conditions, encompassing day and night, indoors and outdoors, and other intricate illumination contexts. 
In contrast, our consistency loss is based on the geometric consistency of the corrected document images captured from different angles. Regardless of the various shooting angles of the input document images, their geometric correction outcomes maintain a point-to-point correspondence in geometry. As such, we can compute the disparities between the estimated warping meshes, which in turn amplifies the robustness and accuracy of the correction.

\smallskip
\textbf{Impact of Deformation Refinement.}
In addition, we investigate the effectiveness of the deformation refinement module. In the ablation study, we eliminate this module from our DRNet, retaining only the deformation initialization module. As illustrated in Table~\ref{aba:aba}, experiments (b) and (c) demonstrate that the refinement module indeed enhances the performance. This improvement can be attributed to the refinement operation, which yields a more precise rectification.

\smallskip
\textbf{Impact of Prediction Target.}
During the training phase, existing methods all incorporate supervision on the output warping mesh and compute the loss, whereas in the inference phase, they rectify the images using the warping mesh produced by the network.
In contrast,
during the training phase, our DRNet integrates supervision on the output rectified images to compute the loss, thus compelling the network to learn the correct warping mesh. During inference, we similarly rectify the distorted images using the output warping mesh.
In the ablation study, we validate our novel supervision manner.

To be consistent with previous methods, we first substitute the image rectification loss $\mathcal{L}_w$ in the Eq.~\eqref{lclb} with the $L_1$ loss applied to the output warping mesh.
As shown in Table~\ref{aba:loss},
experiments (a) and (d) illustrate that our flow-free supervision setting yields better performance.
We hope that this result can inspire more effective supervision manners.

Furthermore,
we investigate the impact of the pixel-wise $L_1$ loss $\mathcal{L}_p$ and the VGG loss $\mathcal{L}_v$ in Eq.~\eqref{eq:lw}.
The results are shown in Table~\ref{aba:loss}.
Experiments (b), (c), and (d) indicate that the performance is optimal when both types of losses are computed.
A plausible explanation is that both the constraints, from the perspectives of low-level pixel alignment and high-level semantic alignment, jointly contribute to the learning of the mapping relationship between deformed document images and distortion-free ones,
thus the rectified image is
both locally and globally accurate and reasonable.

\begin{figure}[t]
  \centering
  \includegraphics[width=0.98\linewidth]{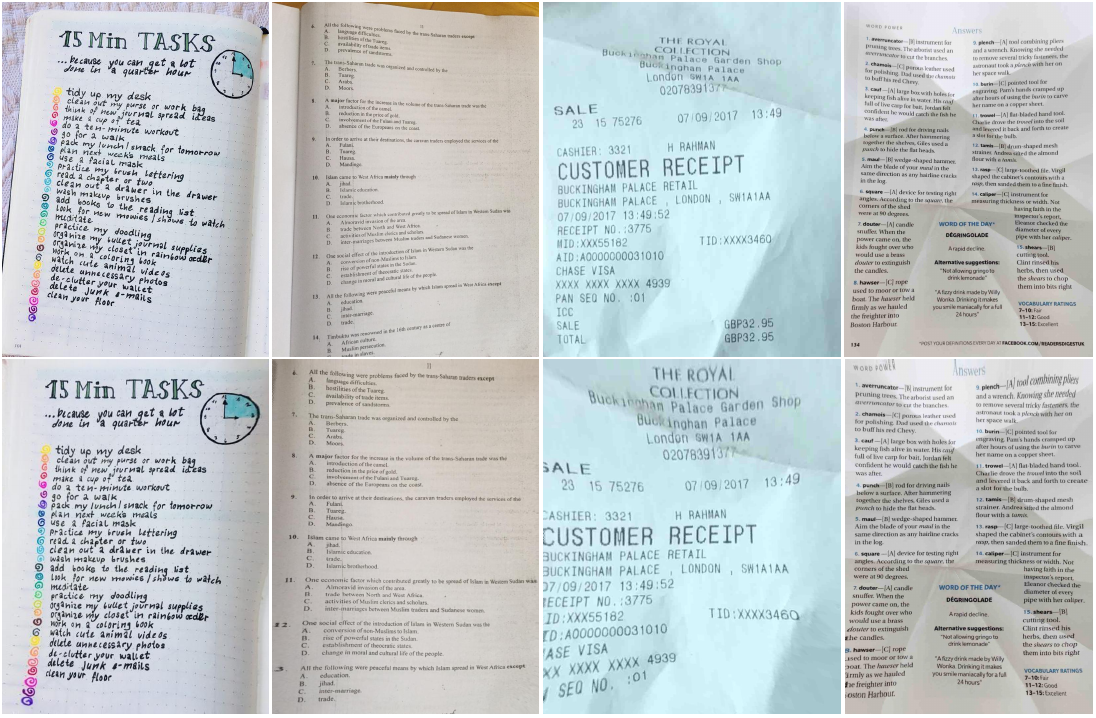}
  \caption{Failure cases of DRNet. The two rows display input distorted and corresponding rectified images, respectively. When the input distorted images have only incomplete (left two columns) or no document
boundaries (right two columns), the rectified images of DRNet remain partially distorted.
  }
  \label{fig:fail}
\end{figure}

\smallskip
\textbf{Impact of Shooting Condition.}
To further verify the robustness of our DRNet, we evaluate the rectification performance in two aspects, including the change of viewpoint and illumination.
The results are shown in Fig.~\ref{fig:view} and Fig.~\ref{fig:illumination}, respectively.
As we can see,
such images are well-rectified,
indicating the robustness of our method.

In Fig.~\ref{fig:showcase},
we present more examples of our method on rectifying various document images, including printed and handwritten cases within different backgrounds.
It can be seen that our method shows outstanding rectification results.
These results demonstrate its strong generalization ability.

\smallskip

\textbf{Impact of the Refinement Times.}
Furthermore, we conduct additional experiments to study the impact of refinement iterations in the DRNet. 
As shown in Table IV, increasing the number of iterations did not lead to further improvements in performance. 
This is due to the increase in the number of refinement modules, 
which in turn increased the model's parameter size, making training more challenging and subsequently degrading performance. 
This phenomenon is also reflected in classic DeepSnake~\cite{peng2020deep} for instance segmentation.

\subsection{Comparison with State-of-the-art Methods}
We evaluate the performance of our DRNet on the DocUNet Benchmark dataset~\cite{8578592} using both quantitative and qualitative evaluation metrics. Table~\ref{tab:t1} presents the comparison of our DRNet with existing state-of-the-art learning-based methods~\cite{8578592,li2019document,xie2020dewarping,9010747,das2021end,xie2021document,xue2022fourier,liudocmae2023}. As evident from the table, DRNet demonstrates superior performance, achieving the highest OCR accuracy with a CER of 0.1649, surpassing the second-best RDGR~\cite{jiang2022revisiting} with a CER of 0.1717. Furthermore, DRNet delivers remarkable performance in distortion rectification, achieving the highest Distortion Level (LD) at 7.42, outperforming Marior~\cite{zhang2022marior} which recorded an LD of 7.44. These quantitative results highlight the effectiveness and competitiveness of our proposed approach in the document image rectification domain, showcasing its ability to tackle challenges in both OCR accuracy and distortion rectification.

As illustrated in Fig.~\ref{fig:viscom}, our proposed DRNet method attains remarkable qualitative rectification performance in comparison to other learning-based techniques. A meticulous examination of the rectified images generated by DRNet reveals a notably reduced degree of distortion, resulting in a more visually coherent and aesthetically pleasing representation. Furthermore, the corrected text lines in the images processed by DRNet display a higher degree of linearity, demonstrating the method's efficacy in rectifying textual content and enhancing readability. This superior qualitative performance underscores the advantages of our DRNet approach in addressing the challenges associated with document image rectification, which ultimately benefits a wide range of downstream applications.

\begin{figure*}[t]
  \centering
  \includegraphics[width=1\linewidth]{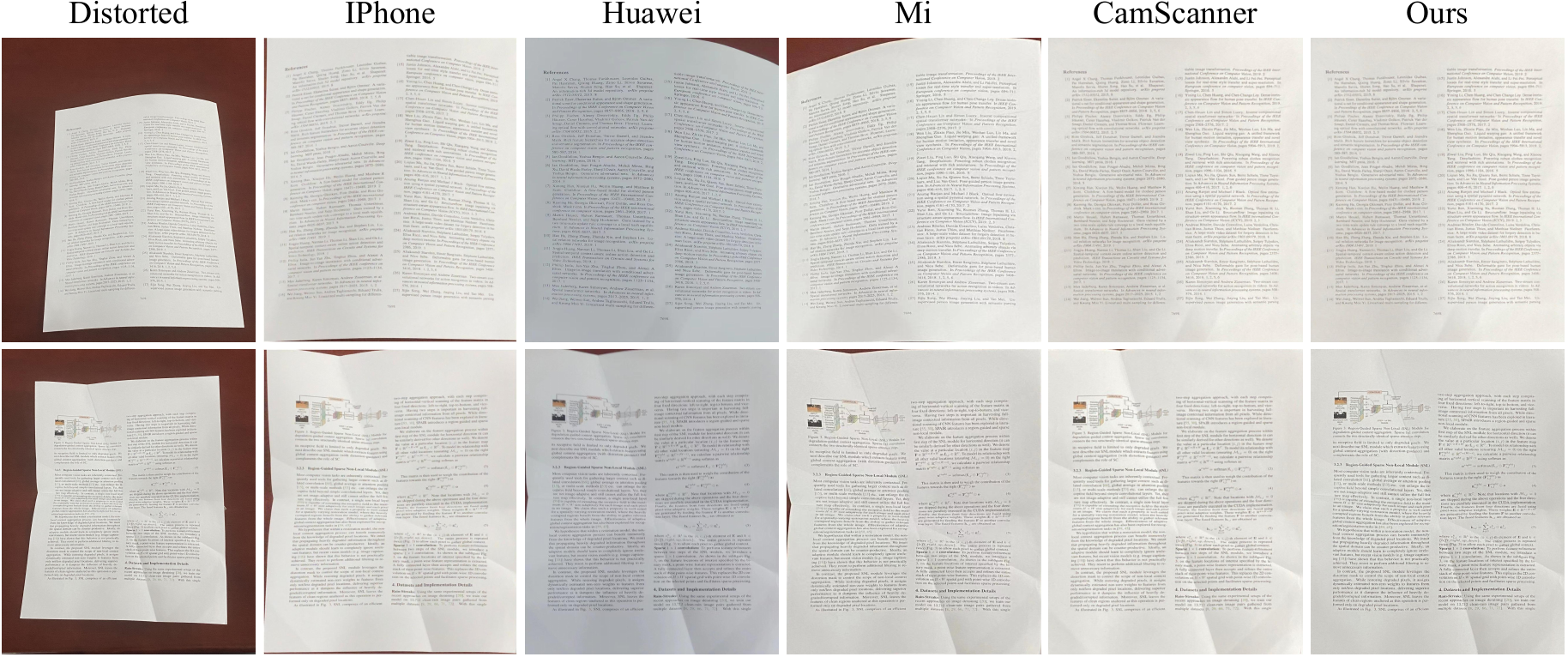}
  \caption{Qualitative comparisons with the prevalent techniques in smartphones. The two rows show input distorted and corresponding rectified images. Different from the prevalent techniques in smartphones, the rectified images of our method reveal less distortion.}
  \label{fig:app}
 
\end{figure*} 

\subsection{Comparison with Industrial Solutions}
The prevailing methodologies employed in smartphone applications often impose a limitation that the scanned document must exhibit a regular quadrilateral shape. They commonly first identify the corner points of the document to determine its spatial boundaries and subsequently apply a perspective transformation to rectify the captured image. Consequently, these approaches are unable to handle scenarios where the scanned document exhibits irregular deformations.

We juxtapose our DRNet against several prevailing industrial solutions, including the CamScanner Application \footnote{https://www.camscanner.com}, the built-in document rectification system of iPhone 11 Pro Max\footnote{https://support.apple.com/kb/SP806}, Huawei Nova 9\footnote{https://consumer.huawei.com/en/phones/nova9}, and Xiaomi 11\footnote{https://www.mi.com/global/product/mi-11}.
As shown in Fig.~\ref{fig:app},
our DRNet showcases the ability to rectify diverse irregular deformations. This capability stems from the predicted warping flow by our DRNet, which utilizes a non-parametric transformation capable of accommodating a wide spectrum of distortions.
Besides,
it is worth noting that CamScanner has recently launched a distortion correction feature. However, its correction performance is slightly inferior to ours, for example, its corrected text lines are not straight enough.
Besides, the ``Reference" image restored by Huawei exhibits excellent recovery on the left half, while the right half appears to collapse. In contrast, our method robustly recovers the entire image. Moreover, it is worth mentioning that the image restored by Huawei might have employed color/text enhancement techniques to make the image more aesthetically pleasing and the text more readable. In our current work, we solely focus on geometric distortion correction, extracting pixels from the distorted image to reconstruct the rectified image.

Note that a significant advantage of traditional projection transformation methods is their lower computational complexity, allowing for convenient deployment on mobile devices.
In contrast, for deep learning-based document image restoration, due to the computational complexity of the dense prediction networks, these solutions are currently deployed on the cloud.

\subsection{Limitation Discussion}
We further discuss the limitations of DRNet.
As shown in Fig.~\ref{fig:fail},
our DRNet tends to underperform when handling document images that have incomplete borders or lack any defined boundaries. 
This is because our DRNet only focuses on the rectification of images with complete visible document boundaries.
We are aware of recent advances in unconstrained document image rectification, such as DocTr++\cite{feng2023deep} and MataDoc~\cite{dai2023matadoc}. In the future, we aim to further our efforts and optimizations in this direction, building upon our DRNet.

\section{Conclusion}
In this work, 
we introduce DRNet,
an innovative deep network for document image rectification. 
Our approach hinges on three core principles: exploiting the intrinsic consistency of document images to guide the learning of distortion rectification, utilizing a coarse-to-fine rectification network to effectively leverage the extracted representations from distorted document images, and proposing an innovative perspective for supervising the learning of rectification networks, which employs undistorted document images for supervision without necessitating warping flow as ground truth. 
Our method, assessed on the demanding DocUNet Benchmark dataset, 
establishes several state-of-the-art scores, 
thereby illustrating its effectiveness and superiority over existing techniques. To further substantiate the efficacy of our approach, we have also conducted an extensive series of ablation experiments.

{
	\bibliographystyle{IEEEtran}
	\bibliography{ref}

@article{li2019document,
	title={Document rectification and illumination correction using a patch-based {CNN}},
	author={Li, Xiaoyu and Zhang, Bo and Liao, Jing and Sander, Pedro V},
	journal={ACM Transactions on Graphics},
	volume={38},
	number={6},
	pages={1--11},
	year={2019}
}

@inproceedings{xie2020dewarping,
	title={Dewarping Document Image by Displacement Flow Estimation with Fully Convolutional Network},
	author={Xie, Guowang and Yin, Fei and Zhang, Xuyao and Liu, Chenglin},
	booktitle={International Workshop on Document Analysis Systems},
	pages={131--144},
	year={2020}
}

@ARTICLE{Shaodi,  
	author={You, Shaodi and Matsushita, Yasuyuki and Sinha, Sudipta and Bou, Yusuke and Ikeuchi, Katsushi},  
	journal={IEEE Transactions on Pattern Analysis and Machine Intelligence},   
	title={Multiview Rectification of Folded Documents},   
	year={2018},  
	volume={40}, 
	number={2}, 
	pages={505-511}
}

@ARTICLE{levenshtein1966binary,
  title={Binary codes capable of correcting deletions, insertions, and reversals},
  author={Levenshtein, Vladimir I},
  journal={Soviet Physics Doklady},
  volume={10},
  number={},
  pages={707--710},
  year={1966}
}

@ARTICLE{1284395,  
	author={Wang, Zhou and  Bovik, Alan C and Sheikh, Hamid R. and Simoncelli, Eero P.},  
	journal={IEEE Transactions on Image Processing},   
	title={Image quality assessment: from error visibility to structural similarity},   
	year={2004},  
	volume={13}, 
	number={4},  
	pages={600-612}, 
}

@INPROCEEDINGS{1292216,  
	author={Wang, Zhou and Simoncelli, Eero P. and Bovik, Alan C.}, 
	booktitle={Proceedings of the Asilomar Conference on Signals, Systems Computers},   
	title={Multiscale structural similarity for image quality assessment},   
	year={2003},  
	volume={2},  
	number={},  
	pages={1398-1402},  
}

@INPROCEEDINGS{4376991,  
	author={Smith, Ray},  
	booktitle={Proceedings of the International Conference on Document Analysis and Recognition},   
	title={An Overview of the Tesseract {OCR} Engine},   
	year={2007},  
	volume={2},  
	number={},  
	pages={629-633},  
}

@ARTICLE{5551153,  
	author={Liu, Ce and Yuen, Jenny and Torralba, Antonio},  
	journal={IEEE Transactions on Pattern Analysis and Machine Intelligence},   
	title={{SIFT} Flow: Dense Correspondence across Scenes and Its Applications},   
	volume={33},  
	number={5},  
	year={2011},  
	pages={978-994},  
}

@inproceedings{loshchilov2017decoupled,
  title={Decoupled Weight Decay Regularization},
  author={I. Loshchilov and F. Hutter},
  booktitle={Proceedings of the International Conference on Learning Representations},
  year={2019}
}

@inproceedings{937649,
      title={Document restoration using {3D} shape},
      author={{M. S. Brown} and {W. B. Seal}},
      booktitle={Proceedings of the IEEE International Conference on Computer Vision},
      pages={9--12},
      year={2001}
}

@ARTICLE{4407722,  author={Zhang, Li and Zhang, Yu and Tan, Chew},  
	journal={IEEE Transactions on Pattern Analysis and Machine Intelligence},   
	title={An Improved Physically-Based Method for Geometric Restoration of Distorted Document Images},  
	year={2008},  
	volume={30},  
	number={4},  
	pages={728-734}
}

@INPROCEEDINGS{6909892,  
	author={Meng, Gaofeng and Wang, Ying and Qu, Shenquan and Xiang, Shiming and Pan, Chunhong},  
	booktitle={Proceedings of the IEEE International Conference on Computer Vision},   
	title={Active Flattening of Curved Document Images via Two Structured Beams},   
	year={2014},  
	volume={},  
	number={},  
	pages={3890-3897}
}

@ARTICLE{4916075,  author={Koo, Hyung Il and Kim, Jinho and Cho, Nam Ik},  
	journal={IEEE Transactions on Image Processing},   
	title={Composition of a Dewarped and Enhanced Document Image From Two View Images},   
	year={2009},  
	volume={18},  
	number={7},  
	pages={1551-1562}
}

@ARTICLE{1561180,  
	author={ Chew Lim Tan and  Li Zhang and  Zheng Zhang and  Tao Xia},  
	journal={IEEE Transactions on Pattern Analysis and Machine Intelligence},   
	title={Restoring warped document images through {3D} shape modeling},   
	year={2006},  
	volume={28},  
	number={2},  
	pages={195-208}
}

@INPROCEEDINGS{6628653,  
	author={He, Yuan and Pan, Pan and Xie, Shufu and Sun, Jun and Naoi, Satoshi},  
	booktitle={Proceedings of the International Conference on Document Analysis and Recognition},   
	title={A Book Dewarping System by Boundary-Based {3D} Surface Reconstruction},   
	year={2013},  
	volume={},  
	number={},  
	pages={403-407}
}

@INPROCEEDINGS{1227630,  
	author={ Huaigu Cao and  Xiaoqing Ding and  Changsong Liu},  
	booktitle={Proceedings of the International Conference on Document Analysis and Recognition},   
	title={Rectifying the bound document image captured by the camera: {A} model based approach},   
	year={2003},  
	volume={1},  
	number={},  
	pages={71-75}
}

@inproceedings{wu2002document,
	title={Document image de-warping for text/graphics recognition},
	author={Wu, Changhua and Agam, Gady},
	booktitle={Proceedings of the Joint IAPR International Workshops on Statistical Techniques in Pattern Recognition and Structural and Syntactic Pattern Recognition},
	pages={348--357},
	year={2002}
}

@INPROCEEDINGS{958227,  
	author={Lavialle, Olivier and Molines, X and Angella, Franck and Baylou, Pierre},  
	booktitle={Proceedings of the International Conference on Image Processing},   
	title={Active contours network to straighten distorted text lines},   
	year={2001},  
	volume={3},  
	number={},  
	pages={748-751}
}

@inproceedings{ronneberger2015u,
	title={{U-Net}: Convolutional networks for biomedical image segmentation},
	author={Ronneberger, Olaf and Fischer, Philipp and Brox, Thomas},
	booktitle={Proceedings of the International Conference on Medical Image Computing and Computer-assisted Intervention},
	pages={234--241},
	year={2015}
}

@inproceedings{markovitz2020can,
	title={Can You Read Me Now? {Content} Aware Rectification Using Angle Supervision},
	author={Markovitz, Amir and Lavi, Inbal and Perel, Or and Mazor, Shai and Litman, Roee},
	booktitle={Proceedings of the European Conference on Computer Vision},
	pages={208--223},
	year={2020}
}

@inproceedings{smith2019super,
  title={Super-convergence: Very fast training of neural networks using large learning rates},
  author={Smith, Leslie N and Topin, Nicholay},
  booktitle={Artificial Intelligence and Machine Learning for Multi-Domain Operations Applications},
  volume={11006},
  pages={1100612},
  year={2019}
}

@article{simonyan2015deep,
  title={Very Deep Convolutional Networks for Large-Scale Image Recognition},
  author={Simonyan, Karen and Zisserman, Andrew},
  journal={arXiv preprint arXiv:1409.1556},
  year={2014}
}

@inproceedings{feng2021doctr,
  title={{DocTr}: Document Image Transformer for Geometric Unwarping and Illumination Correction},
  author={Feng, Hao and Wang, Yuechen and Zhou, Wengang and Deng, Jiajun and Li, Houqiang},
  booktitle={Proceedings of the ACM International Conference on Multimedia},
  pages={273--281},
  year={2021}
}

@article{liu2020geometric,
	title={Geometric rectification of document images using adversarial gated unwarping network},
	author={Liu, Xiyan and Meng, Gaofeng and Fan, Bin and Xiang, Shiming and Pan, Chunhong},
	journal={Pattern Recognition},
	volume={108},
	pages={107576},
	year={2020}
}

@inproceedings{das2021end,
  title={End-to-end Piece-wise Unwarping of Document Images},
  author={Das, Sagnik and Singh, Kunwar Yashraj and Wu, Jon and Bas, Erhan and Mahadevan, Vijay and Bhotika, Rahul and Samaras, Dimitris},
  booktitle={Proceedings of the IEEE International Conference on Computer Vision},
  pages={4268--4277},
  year={2021}
}

@inproceedings{yamashita2004shape,
	title={Shape reconstruction and image restoration for non-flat surfaces of documents with a stereo vision system},
	author={Yamashita, Atsushi and Kawarago, Atsushi and Kaneko, Toru and Miura, Kenjiro T},
	booktitle={Proceedings of the International Conference on Pattern Recognition},
	volume={1},
	pages={482--485},
	year={2004}
}

@article{brown2004image,
	title={Image restoration of arbitrarily warped documents},
	author={Brown, Michael S and Seales, W Brent},
	journal={IEEE Transactions on Pattern Analysis and Machine Intelligence},
	volume={26},
	number={10},
	pages={1295--1306},
	year={2004}
}

@inproceedings{tsoi2007multi,
	title={Multi-view document rectification using boundary},
	author={Tsoi, Yau-Chat and Brown, Michael S},
	booktitle={Proceedings of the IEEE Conference on Computer Vision and Pattern Recognition},
	pages={1--8},
	year={2007}
}

@article{lowe2004distinctive,
	title={Distinctive image features from scale-invariant keypoints},
	author={Lowe, David G},
	journal={International Journal of Computer Vision},
	volume={60},
	number={2},
	pages={91--110},
	year={2004}
}

@article{liang2008geometric,
	title={Geometric rectification of camera-captured document images},
	author={Liang, Jian and DeMenthon, Daniel and Doermann, David},
	journal={IEEE Transactions on Pattern Analysis and Machine Intelligence},
	volume={30},
	number={4},
	pages={591--605},
	year={2008}
}

@article{wada1997shape,
  title={Shape from shading with interreflections under a proximal light source: Distortion-free copying of an unfolded book},
  author={Wada, Toshikazu and Ukida, Hiroyuki and Matsuyama, Takashi},
  journal={International Journal of Computer Vision},
  volume={24},
  number={2},
  pages={125--135},
  year={1997}
}

@article{meng2011metric,
	title={Metric rectification of curved document images},
	author={Meng, Gaofeng and Pan, Chunhong and Xiang, Shiming and Duan, Jiangyong and Zheng, Nanning},
	journal={IEEE Transactions on Pattern Analysis and Machine Intelligence},
	volume={34},
	number={4},
	pages={707--722},
	year={2011}
}

@inproceedings{teed2020raft,
  title={{RAFT}: {Recurrent} all-pairs field transforms for optical flow},
  author={Teed, Zachary and Deng, Jia},
  booktitle={Proceedings of the European Conference on Computer Vision},
  pages={402--419},
  year={2020}
}

@inproceedings{xie2021document,
  title={Document Dewarping with Control Points},
  author={Xie, Guo-Wang and Yin, Fei and Zhang, Xu-Yao and Liu, Cheng-Lin},
  booktitle={Proceedings of the International Conference on Document Analysis and Recognition},
  pages={466--480},
  year={2021}
}

@inproceedings{xue2022fourier,
  title={Fourier Document Restoration for Robust Document Dewarping and Recognition},
  author={Xue, Chuhui and Tian, Zichen and Zhan, Fangneng and Lu, Shijian and Bai, Song},
  booktitle={Proceedings of the IEEE Conference on Computer Vision and Pattern Recognition},
  pages={4573--4582},
  year={2022}
}

@inproceedings{feng2022geometric,
	title={Geometric Representation Learning for Document Image Rectification},
    author={Feng, Hao and Zhou, Wengang and Deng, Jiajun and Wang, Yuechen and Li, Houqiang},
	booktitle={Proceedings of the European Conference on Computer Vision},
	year={2022}
}

@inproceedings{tsoi2004geometric,
  title={Geometric and shading correction for images of printed materials: a unified approach using boundary},
  author={Tsoi, YauChat and Brown, Michael S},
  booktitle={Proceedings of the IEEE Conference on Computer Vision and Pattern Recognition},
  volume={1},
  pages={I--I},
  year={2004}
}

@INPROCEEDINGS{8578592,  
	author={Ma, Ke and Shu, Zhixin and Bai, Xue and Wang, Jue and Samaras, Dimitris},  
	booktitle={Proceedings of the IEEE International Conference on Computer Vision},   
	title={{DocUNet}: {Document} Image Unwarping via a Stacked {U-Net}},   
	year={2018},  
	volume={},  
	number={},  
	pages={4700-4709}
}

@INPROCEEDINGS{9010747,  
	author={Das, Sagnik and Ma, Ke and Shu, Zhixin and Samaras, Dimitris and Shilkrot, Roy},  
	booktitle={Proceedings of the International Conference on Computer Vision},   
	title={{DewarpNet}: {Single}-Image Document Unwarping With Stacked {3D} and {2D} Regression Networks},  
	year={2019},  
	volume={},  
	number={},  
	pages={131-140}
}

@INPROCEEDINGS{Vaswani2017AttentionIA,
  title={Attention is All you Need},
  author={Ashish Vaswani and Noam Shazeer and Niki Parmar and Jakob Uszkoreit and Llion Jones and Aidan N. Gomez and Lukasz Kaiser and Illia Polosukhin},
  booktitle={Proceedings of the Neural Information Processing Systems},
  year={2017},
  pages={6000--6010},
}

@inproceedings{jiang2022revisiting,
  title={Revisiting Document Image Dewarping by Grid Regularization},
  author={Jiang, Xiangwei and Long, Rujiao and Xue, Nan and Yang, Zhibo and Yao, Cong and Xia, Gui-Song},
  booktitle={Proceedings of the IEEE Conference on Computer Vision and Pattern Recognition},
  pages={4543--4552},
  year={2022}
}

@inproceedings{ma2022learning,
  title={Learning From Documents in the Wild to Improve Document Unwarping},
  author={Ma, Ke and Das, Sagnik and Shu, Zhixin and Samaras, Dimitris},
  booktitle={Proceedings of the ACM SIGGRAPH Conference},
  pages={1--9},
  year={2022}
}

@inproceedings{ciardiello1988experimental,
  title={An experimental system for office document handling and text recognition},
  author={Ciardiello, G and Scafuro, G and Degrandi, MT and Spada, MR and Roccotelli, MP},
  booktitle={Proceedings of the International Conference on Pattern Recognition},
  pages={739--743},
  year={1988}
}

@article{salton1991developments,
  title={Developments in automatic text retrieval},
  author={Salton, Gerard},
  journal={Science},
  volume={253},
  number={5023},
  pages={974--980},
  year={1991},
}

@inproceedings{wang2011end,
  title={End-to-end scene text recognition},
  author={Wang, Kai and Babenko, Boris and Belongie, Serge},
  booktitle={Proceedings of the International Conference on Computer Vision},
  pages={1457--1464},
  year={2011}
}

@inproceedings{yuan2019interpreting,
  title={Interpreting deep models for text analysis via optimization and regularization methods},
  author={Yuan, Hao and Chen, Yongjun and Hu, Xia and Ji, Shuiwang},
  booktitle={Proceedings of the AAAI Conference on Artificial Intelligence},
  volume={33},
  number={01},
  pages={5717--5724},
  year={2019}
}

@inproceedings{kim2022ocr,
  title={{OCR}-Free Document Understanding Transformer},
  author={Kim, Geewook and Hong, Teakgyu and Yim, Moonbin and Nam, JeongYeon and Park, Jinyoung and Yim, Jinyeong and Hwang, Wonseok and Yun, Sangdoo and Han, Dongyoon and Park, Seunghyun},
  booktitle={Proceedings of the European Conference on Computer Vision},
  pages={498--517},
  year={2022}
}

@inproceedings{lat2018enhancing,
  title={Enhancing {OCR} accuracy with super resolution},
  author={Lat, Ankit and Jawahar, CV},
  booktitle={Proceedings of the International Conference on Pattern Recognition},
  pages={3162--3167},
  year={2018}
}

@inproceedings{mathew2021docvqa,
  title={{DocVQA}: A dataset for {VQA} on document images},
  author={Mathew, Minesh and Karatzas, Dimosthenis and Jawahar, CV},
  booktitle={Proceedings of the IEEE Winter Conference on Applications of Computer Vision},
  pages={2200--2209},
  year={2021}
}

@inproceedings{long2015fully,
  title={Fully convolutional networks for semantic segmentation},
  author={Long, Jonathan and Shelhamer, Evan and Darrell, Trevor},
  booktitle={Proceedings of the IEEE Conference on Computer Vision and Pattern Recognition},
  pages={3431--3440},
  year={2015}
}

@inproceedings{zhang2022marior,
  title={Marior: {Margin} Removal and Iterative Content Rectification for Document Dewarping in the Wild},
  author={Zhang, Jiaxin and Luo, Canjie and Jin, Lianwen and Guo, Fengjun and Ding, Kai},
  booktitle={Proceedings of the ACM International Conference on Multimedia},
  pages={2805--2815},
  year={2022}
}

@inproceedings{meng2018exploiting,
  title={Exploiting vector fields for geometric rectification of distorted document images},
  author={Meng, Gaofeng and Su, Yuanqi and Wu, Ying and Xiang, Shiming and Pan, Chunhong},
  booktitle={Proceedings of the European Conference on Computer Vision},
  pages={172--187},
  year={2018}
}

@article{meijering2002chronology,
  title={A chronology of interpolation: {From} ancient astronomy to modern signal and image processing},
  author={Meijering, Erik},
  journal={Proceedings of the IEEE},
  volume={90},
  number={3},
  pages={319--342},
  year={2002}
}

@article{nie2012beyond,
  title={Beyond text {QA}: multimedia answer generation by harvesting web information},
  author={Nie, Liqiang and Wang, Meng and Gao, Yue and Zha, Zheng-Jun and Chua, Tat-Seng},
  journal={IEEE Transactions on Multimedia},
  volume={15},
  number={2},
  pages={426--441},
  year={2012}
}

@article{yang2008harmonizing,
  title={Harmonizing hierarchical manifolds for multimedia document semantics understanding and cross-media retrieval},
  author={Yang, Yi and Zhuang, Yue-Ting and Wu, Fei and Pan, Yun-He},
  journal={IEEE Transactions on Multimedia},
  volume={10},
  number={3},
  pages={437--446},
  year={2008}
}

@article{zhang2022multimodal,
  title={Multimodal pre-training based on graph attention network for document understanding},
  author={Zhang, Zhenrong and Ma, Jiefeng and Du, Jun and Wang, Licheng and Zhang, Jianshu},
  journal={IEEE Transactions on Multimedia},
  year={2022}
}

@article{luo2022geometric,
  title={Geometric Rectification of Creased Document Images based on Isometric Mapping},
  author={Luo, Dong and Bo, Pengbo},
  journal={arXiv preprint arXiv:2212.08365},
  year={2022}
}

@inproceedings{das2020intrinsic,
  title={Intrinsic Decomposition of Document Images In-the-Wild},
  author={Das, S and Sial, H Ma and Baldrich, R and Vanrell, M and Samaras, D},
  booktitle={Proceedings of the British Machine Vision Conference},
  year={2020}
}

@article{peng2022recognition,
  title={Recognition of handwritten Chinese text by segmentation: {A} segment-annotation-free approach},
  author={Peng, Dezhi and Jin, Lianwen and Ma, Weihong and Xie, Canyu and Zhang, Hesuo and Zhu, Shenggao and Li, Jing},
  journal={IEEE Transactions on Multimedia},
  year={2022}
}

@article{liu2023end,
  title={End-to-end learning of representations for instance-level document image retrieval},
  author={Liu, Li and Lu, Yue and Suen, Ching Y},
  journal={Applied Soft Computing},
  pages={110136},
  year={2023}
}

@inproceedings{liudocmae2023,
  title={{DocMAE}: Document Image Rectification via
Self-supervised Representation Learning},
  author={Liu, Shaokai and Feng, Hao and Zhou, Wengang and Li, Houqiang and Liu, Cong and Wu, Feng},
  booktitle={Proceedings of the International Conference on Multimedia and Expo},
  year={2023}
}

@article{verhoeven2023neural,
  title={Neural Document Unwarping using Coupled Grids},
  author={Verhoeven, Floor and Magne, Tanguy and Sorkine-Hornung, Olga},
  journal={arXiv preprint arXiv:2302.02887},
  year={2023}
}

@inproceedings{he2022masked,
  title={Masked autoencoders are scalable vision learners},
  author={He, Kaiming and Chen, Xinlei and Xie, Saining and Li, Yanghao and Doll{\'a}r, Piotr and Girshick, Ross},
  booktitle={Proceedings of the IEEE Conference on Computer Vision and Pattern Recognition},
  pages={16000--16009},
  year={2022}
}

@inproceedings{ViT,
  title={An Image is Worth 16x16 Words: Transformers for Image Recognition at Scale},
  author={Dosovitskiy, Alexey and Beyer, Lucas and Kolesnikov, Alexander and Weissenborn, Dirk and Zhai, Xiaohua and Unterthiner, Thomas and Dehghani, Mostafa and Minderer, Matthias and Heigold, Georg and Gelly, Sylvain and others},
  booktitle={Proceedings of the International Conference on Learning Representations},
  year={2021}
}

@inproceedings{jaderberg2015spatial,
	title={Spatial transformer networks},
	author={Jaderberg, Max and Simonyan, Karen and Zisserman, Andrew and others},
	booktitle={Proceedings of the Neural Information Processing Systems},
	pages={2017--2025},
	year={2015}
}

@article{feng2023deep,
  title={Deep Unrestricted Document Image Rectification},
  author={Feng, Hao and Liu, Shaokai and Deng, Jiajun and Zhou, Wengang and Li, Houqiang},
  journal={arXiv preprint arXiv:2304.08796},
  year={2023}
}

@article{liu2023visual,
  title={Visual instruction tuning},
  author={Liu, Haotian and Li, Chunyuan and Wu, Qingyang and Lee, Yong Jae},
  journal={arXiv preprint arXiv:2304.08485},
  year={2023}
}

@article{li2022emu,
  title={Emu: Effective multi-hot encoding net for lightweight scene text recognition with a large character set},
  author={Li, Bingcong and Tang, Xin and Qi, Xianbiao and Chen, Yihao and Li, Chun-Guang and Xiao, Rong},
  journal={IEEE Transactions on Circuits and Systems for Video Technology},
  volume={32},
  number={8},
  pages={5374--5385},
  year={2022},
  publisher={IEEE}
}

@article{yu2008semantic,
  title={Semantic subspace projection and its applications in image retrieval},
  author={Yu, Jie and Tian, Qi},
  journal={IEEE Transactions on Circuits and Systems for Video Technology},
  volume={18},
  number={4},
  pages={544--548},
  year={2008},
  publisher={IEEE}
}

@article{zhou2022hrinversion,
  title={HRInversion: High-Resolution GAN Inversion for Cross-Domain Image Synthesis},
  author={Zhou, Peng and Xie, Lingxi and Ni, Bingbing and Liu, Lin and Tian, Qi},
  journal={IEEE Transactions on Circuits and Systems for Video Technology},
  year={2022},
  publisher={IEEE}
}

@article{qin2020u2,
  title={{U2-Net}: {Going} deeper with nested {U-structure} for salient object detection},
  author={Qin, Xuebin and Zhang, Zichen and Huang, Chenyang and Dehghan, Masood and Zaiane, Osmar R and Jagersand, Martin},
  journal={Pattern recognition},
  volume={106},
  pages={107404},
  year={2020}
}

@inproceedings{ren2019progressive,
  title={Progressive image deraining networks: A better and simpler baseline},
  author={Ren, Dongwei and Zuo, Wangmeng and Hu, Qinghua and Zhu, Pengfei and Meng, Deyu},
  booktitle={Proceedings of the IEEE Conference on Computer Vision and Pattern Recognition},
  pages={3937--3946},
  year={2019}
}

@article{buades2005review,
  title={A review of image denoising algorithms, with a new one},
  author={Buades, Antoni and Coll, Bartomeu and Morel, Jean-Michel},
  journal={Multiscale modeling \& simulation},
  volume={4},
  number={2},
  pages={490--530},
  year={2005}
}

@inproceedings{hertlein2023template,
  title={Template-Guided Illumination Correction for Document Images with Imperfect Geometric Reconstruction},
  author={Hertlein, Felix and Naumann, Alexander},
  booktitle={Proceedings of the IEEE International Conference on Computer Vision},
  pages={904--913},
  year={2023}
}

@inproceedings{li2023foreground,
  title={Foreground and Text-lines Aware Document Image Rectification},
  author={Li, Heng and Wu, Xiangping and Chen, Qingcai and Xiang, Qianjin},
  booktitle={Proceedings of the IEEE International Conference on Computer Vision},
  pages={19574--19583},
  year={2023}
}

@inproceedings{peng2020deep,
  title={Deep snake for real-time instance segmentation},
  author={Peng, Sida and Jiang, Wen and Pi, Huaijin and Li, Xiuli and Bao, Hujun and Zhou, Xiaowei},
  booktitle={Proceedings of the IEEE International Conference on Computer Vision},
  pages={8533--8542},
  year={2020}
}

@article{dai2023matadoc,
  title={{MataDoc}: Margin and Text Aware Document Dewarping for Arbitrary Boundary},
  author={Dai, Beiya and Xie, Qunyi and Li, Yulin and Qin, Xiameng and Zhang, Chengquan and Yao, Kun and Han, Junyu and others},
  journal={arXiv preprint arXiv:2307.12571},
  year={2023}
}

@article{feng2021docscanner,
  title={{DocScanner}: {Robust} document image rectification with progressive learning},
  author={Feng, Hao and Zhou, Wengang and Deng, Jiajun and Tian, Qi and Li, Houqiang},
  journal={arXiv preprint arXiv:2110.14968},
  year={2021}
}
}

\end{document}